\def\year{2020}\relax
\documentclass[letterpaper]{article} 
\usepackage{aaai/aaai20}  
\usepackage{times}  
\usepackage{helvet} 
\usepackage{courier}  
\usepackage[hyphens]{url}  
\usepackage{graphicx} 
\usepackage{dirtytalk}
\usepackage{graphicx}  
\usepackage{algorithm2e}
\usepackage{subcaption}
\title{Amnesiac Machine Learning}

\author{Laura Graves, Vineel Nagisetty, Vijay Ganesh\\ 
University of Waterloo\\ 
laura.graves@uwaterloo.ca, vineel.nagisetty@uwaterloo.ca, vganesh@uwaterloo.ca
}

\begin{document}

\maketitle

\begin{abstract}
The {\it Right to be Forgotten} is part of the recently enacted General Data Protection Regulation (GDPR) law that affects any data holder that has data on European Union residents. It gives EU residents the ability to request deletion of their personal data, including training records used to train machine learning models. Unfortunately, Deep Neural Network models are vulnerable to information leaking attacks such as model inversion attacks which extract class information from a trained model and membership inference attacks which determine the presence of an example in a model's training data. If a malicious party can mount an attack and learn private information that was meant to be removed, then it implies that the model owner has not properly protected their user's rights and their models may not be compliant with the GDPR law. In this paper, we present two efficient methods that address this question of how a model owner or data holder may delete personal data from models in such a way that they may not be vulnerable to model inversion and membership inference attacks while maintaining model efficacy. We start by presenting a real-world threat model that shows that simply removing training data is insufficient to protect users. We follow that up with two data removal methods, namely Unlearning and Amnesiac Unlearning, that enable model owners to protect themselves against such attacks while being compliant with regulations. We provide extensive empirical analysis that show that these methods are indeed efficient, safe to apply, effectively remove learned information about sensitive data from trained models while maintaining model efficacy.
\end{abstract}

\section{Introduction}

In 2016 the European Union (EU) established the {\it General Data Protect Regulation} (GDPR) which is intended to provide individuals in EU nations control over their personal data. This includes regulations for businesses that handle personal data, requiring them to provide safeguards to protect data and use the highest possible privacy settings by default~\cite{GDPR}. In particular, article 17 of the GDPR gives individuals the right to be forgotten and states that \say{... (businesses) have the obligation to erase personal data without undue delay}~\cite{GDPR}. Individuals who invoke this right must have their personal data removed from data records that companies, governments, or other entities hold. 

In light of the GDPR law, the existence of attacks on neural networks that leak information about the data they were trained on, such as the Model Inversion Attack~\cite{Fredrikson2015ModelIA} and the Membership Inference Attack~\cite{membership4}, present a problem for companies or researchers that use personal data to train neural network models. If an individual's data has been used to train a model and that individual subsequently invokes the right to be forgotten, simply deleting the training data is insufficient to protect the individual's privacy. The reason is that neural networks, whose training set contains a specific datum, are vulnerable to leakage of that datum via the above-mentioned attacks. Although this problem is recent, legal scholars have underlined that the potential cost due to GDPR can be significant and potentially crippling to businesses and entities~\cite{takingaipersonally},\cite{villaronga2018humans}. If a malicious entity or attacker can attack a trained model and learn private information about an individual who has invoked the right to be forgotten, the model owner can be held liable since he or she has not taken the appropriate steps to protect the individual.

In this paper, we address the question of efficiently removing learned data from a trained neural network without harming the performance of the network. We call our approach amnesiac unlearning and provide extensive empirical analysis to highlight its strengths as well as understand its weaknesses. The obvious solution to this problem is to entirely discard the trained model and train a new one from scratch. Unfortunately, training machine learning models is known to be an expensive and time-consuming proposition. Model owners are likely to be disinclined to continually train new models to be compliant with regulations every time an individual invokes their right to be forgotten. Hence, it is in their best interest to have efficient deletion methods to remove learned data that ensure compliance with regulations and retain the fidelity of their model. Humerick~\cite{takingaipersonally} highlights the importance of having a method that effectively removes learned data, as well as the potential economic harm that could be caused by requiring model owners to continually replace trained models.

\vspace{0.1cm}
\noindent{\bf Problem Statement:} In short, our work focuses on the following problem: {\it how can learned data be safely and efficiently removed from a trained neural network without harming the performance of the network and that is not vulnerable to state-of-the-art data leakage attacks?}

\subsection{Contributions}

In this paper, we make the following contributions:

\begin{itemize}
    \item We address the problem of how to make neural networks compliant with the right to be forgotten policy of the GDPR law via effective ways of removing learned data. In this context, we introduce two methods for efficient removal of learned data, namely {\it Unlearning} and {\it Amnesiac Unlearning}. Unlearning is a training-time method that uses relatively little time to remove learned data from a network. By contrast, amnesiac unlearning is a single-step method that is effective for laser-focused removal of small segments of data. We compare these methods against the naive method of retraining a model without the sensitive data. We show that these methods are not only very efficient, but effective in removing traces of the data that could be leaked through state-of-the-art attacks, and further do not harm the performance of the neural network in any other way.
    
    \item We provide a detailed empirical evaluation of the efficacy of our methods along several vectors, including protection against data leaks, efficiency, and model performance. Specifically, we evaluate protection of the proposed methods using state-of-the-art privacy leaking attacks, such as model inversion and membership inference, and show that amnesiac unlearning is the most effective when compared to other methods. As part of this evaluation, we present a modified version of the Model Inversion Attack~\cite{Fredrikson2015ModelIA} that is effective even against complex convolutional networks, a class of network previously considered impervious to such attacks~\cite{inversiongan}. Further, we show that both unlearning and amnesiac unlearning are very efficient to apply and cost very little to model owners. Finally, we evaluate the performance of neural network models after the application of these deletion methods, and show that these methods have virtually no impact on model performance on data that is unrelated to the deleted data.
\end{itemize}

\section{Related Work}

Machine learning models have been shown to leak information about the data they've been trained on~\cite{membership2,membership4,184489,Fredrikson2015ModelIA,inversiongan}. Specifically, two main kinds of information leaking attacks have been studied: {\it membership inference attacks} that leak information about the presence of specific records in the training data~\cite{membership5} and {\it model inversion attacks} that leak class information~\cite{Fredrikson2015ModelIA},\cite{inversiongan}. 
Membership inference attacks determine whether a particular record was present in the training data for a model. This attack was first presented in 2008~\cite{membership2} and was formalized in 2015~\cite{membership1}. Since then, considerable work has been done on membership inference attacks and defense mechanisms against such membership inferences~\cite{membership3},\cite{membership4},\cite{membership5}. Property inference attacks are a subset of membership inference attacks that determine a general property of the training data, such as the ratio of training examples in each class~\cite{propertyinference}. Model inversion attacks, introduced by Fredrikson et al. in 2014~\cite{184489} and expanded to vision tasks in 2015~\cite{Fredrikson2015ModelIA}, have been shown to recreate instances of records from trained ML models. Given white-box access to a trained model, examples of target classes or points near a regression value can be recreated. In this paper, we utilize both state-of-the-art membership inference attacks and model inversion attacks to evaluate how likely a model is to leak private data.

There is an abundance of literature on differential privacy which provides an upper bound on the amount of information that can be leaked from each individual data record~\cite{dp1},\cite{dp2},\cite{dp3},\cite{dp4}. Cummings and Desai address the need for training machine learning models in a differentially private manner to comply with GDPR~\cite{cummings2018role}. However, differential privacy methods do not allow learned data to be forgotten. Any of the methods presented in this paper can be combined with differential privacy methods if this type of privacy guarantee is needed.

Other recent research has touched on the issue of removing learned properties or data from machine learning models. Ginart et al.~\cite{makingaiforgetyou} devised a notion they term {\it removal efficiency} and give two algorithms for efficiently removing specific data points from $k$-means clustering models. Guo et al.~\cite{certifiedremoval} defined an approach they term {\it certified removal} that was evaluated for linear classification models. In this system, a model is trained on a dataset including sensitive data, and then the sensitive data is removed in some way from the model. A different model is trained on the same dataset without the sensitive data, and removal algorithms are evaluated based on the difference between the models. However, these approaches are unlikely to work in the case of DNNs, which are more opaque and difficult to analyze. The work presented in Ginart et al.~\cite{makingaiforgetyou} and Guo et al.~\cite{certifiedremoval} are both focused on such different machine learning methods that it is impossible to make a direct comparison with our work.

Recently, Bourtoule et al.~\cite{machineunlearning} introduced a method for dealing with individual data removal requests. They proposed SISA training, a method consisting of an aggregate model made of multiple models trained on disjoint partitions of the data. Because of this segmentation, when requests for removal are made the model owner can retrain only the effected sub-models instead of having to retrain everything. By contrast, our methods focus on deep learning models that have already been trained and are independent of neural network architecture. Further, we have no requirement to have the model be an aggregate of weak learners. 

\section{Proposed Methods}

In this section, we explain our proposed methods for removing learned data from a neural network model.


\subsection{Naive Retraining}

As a baseline, we consider the naive method of simply removing the sensitive data from the dataset and retraining the model anew without the sensitive data. The principle of catastrophic interference~\cite{sharkey1995analysis} tells us that when presented with new data, there is some probability of an artificial neural network losing previously learned information. Over time a network originally trained on a dataset $D$ and then trained on the same dataset without some subset of that data $D \setminus S$ has a high probability of losing information about $S$. Although this has a high probability of causing the neural network to forget the desired information, it gives no information or assurances about how long it could take. However, it provides a useful baseline because simply removing the data and retraining the network is a simple solution and, if useful, provides model owners with an easy method to cause networks to forget sensitive data. Unfortunately, our evaluation shows that naive retraining takes far too long to be considered as a practical solution and even a large amount of naive retraining does not prevent data leaks.
\subsection{Unlearning}

The goal of unlearning is to muddy the model's understanding of the sensitive data to the point that it is unable to retain any meaningful information about that data. To achieve this, we relabel the sensitive data with randomly selected incorrect labels and then retrain the network for some iterations on the modified dataset. Unlearning for an entire class is achieved by replacing the label for each example in that class with a randomly selected incorrect label, whereas unlearning for a select set of examples is achieved by removing those examples and inserting a small number of copies of each of them with randomly selected incorrect labels. This relabeling is computationally inexpensive and our evaluation shows that this method is effective with only a very small number of training iterations on the modified dataset.

One possible risk of unlearning is that the data holder must maintain a copy of the sensitive data during the unlearning process, which may potentially have legal significance. Fortunately, the right to be forgotten gives data holders up to a month to remove the data~\cite{ico}, which makes this method easily fall within permissible bounds.
\subsection{Amnesiac Unlearning}

Amnesiac unlearning involves selectively undoing the learning steps that involved the sensitive data. During training, the model owner keeps a list of which examples appeared in which batches as well as the parameter updates from each batch. When a data removal request comes, the model owner undoes the parameter updates from only the batches containing sensitive data. As long as the number of batches effected is small, the effect of undoing those sections of learning is also small. One significant advantage of this method is that it can very effectively remove the learning from a single example with minimal impact on the rest of the learned model, which is something other techniques have struggled with. Further, this method is very time efficient, and our evaluation shows that as long as the amount of data to remove is small the efficacy of the model remains unharmed.

To better understand amnesiac unlearning, we start by revisiting model training with a fresh perspective. We can view model training as a series of parameter updates to the initial model parameters (that, in the case of neural networks, are randomly initialized). We begin from initial model parameters $\theta_{initial}$. Model $M$ is then trained for $E$ epochs each consisting of $B$ batches, and the parameters are updated after each batch by an amount $\Delta_{\theta_{e,b}}$. The learned model parameters can then be expressed as:

\begin{center}
    $\theta_M = \theta_{initial} + \sum\limits_{e=1}^E \sum\limits_{b=1}^B \Delta_{\theta_{e,b}}$
\end{center}


During training, we keep a list $SB$ of which batches contained the sensitive data. This can be in the form of an index of batches for each example in the training data, an index of batches for each class, or any other form desired. We also must maintain the model parameter updates from each batch that contained sensitive data. A comprehensive approach would be to keep a record of each batch update. However, if the data holder is only concerned about possible potential removal of a subset of data, they need only keep the parameter updates from batches containing that data.

Once training is completed, we can produce a protected model $M'$ using amnesiac unlearning as follows: we simply remove the parameter updates from each batch $sb \in SB$ (the list of sensitive data batches) from the learned parameters $\theta_M$.

\begin{center}
    $\theta_{M'} = \theta_{initial} + \sum\limits_{e=1}^E \sum\limits_{b=1}^B \Delta_{\theta_{e,b}} - \sum\limits_{sb=1}^{SB} \Delta_{\theta_{sb}} = \theta_M - \sum\limits_{sb=1}^{SB} \Delta_{\theta_{sb}}$
\end{center}

We note that the parameter update at each step $\Delta_{\theta_{e,b}}$ depends on the current parameters when that learning step is taken. Due to this, a model trained on three batches $b_0$, $b_1$, and $b_2$ that then has the parameter updates from $b_1$ undone will not in fact be the same as a model (with the same initial parameters) trained directly on $b_0$ and $b_2$, because the parameter updates from learning on $b_2$ will be different. 

When the number of batches in $SB$ is low, the difference is between $\theta_M$ and $\theta_{M'}$ is comparatively low, and the concomitant impact on the efficacy of the model is also low. In this way, this method provides a way of laser-focused removal of sensitive data, i.e., the learned data from a single record can be removed from the model with minimal effect on the rest of the model. This method is particularly well-suited for situations where the privacy of an individual with a single record in the dataset needs to be protected. When the number of batches in $SB$ is higher, the model needs a small amount of fine-tuning after the amnesiac unlearning step to regain performance.

One potential downside to this method is the large storage space required to keep a set of parameter update values from each batch. While this cost can be quite large, especially for state-of-the-art large models, the low cost of mass storage means this cost is usually less than the cost of retraining a full model from scratch. Model owners concerned about this storage cost may be better off using a method such as unlearning that doesn't require this storage overhead.
\subsection{Threat Model}

We use two state-of-the-art attack methods to evaluate how much data can be leaked, namely, model inversion and membership inference attacks. The adversary attempts to gain information about either the class (through model inversion attacks) or about the presence of a specific record or set of records in the training data (through membership inference attacks). Leaking class information could provide an attacker with generalized information they should not have. For example, if the adversary learns that the class of individuals that are at risk for cancer is strongly generalized to be older men, it could potentially violate privacy guidelines that protect the older men represented in the data set. Leaking membership information could also pose a significant privacy threat. For example, if an adversary learns that an individual was represented in a dataset used to train a model to determine bankruptcy risk, they could determine the individual's private bankruptcy history.

We consider the scenario where the adversary has white-box access to the currently published version of the model, but does not have access to any previously published versions. In the case of the model inversion attack the adversary does not have information about what each class represents, and we consider an attack successful if the adversary is able to glean information about what the class represents through model inversion. In the case of the membership inference attack the adversary has access to data from a similar distribution to the one used to train the target model. The data used to train the target model and the data available to the adversary can contain duplicate records, but does not have to.


\section{Empirical Evaluation}

We conduct extensive experiments to evaluate the efficacy of our unlearning methods in comparison to the naive retraining method. First, we conduct an evaluation of the model accuracy during our methods, looking at both the model accuracy of the target data (that we remove) as well as the accuracy on the non-target data (we sometimes refer to this as model performance). Second, we perform model inversion attacks against the models at different stages of the data removal process to show how data can be retained and subsequently leaked for a considerable time period after model accuracy starts to degrade and to evaluate the efficacy of the data removal methods to stop this data leaking. Next, we perform membership inference attacks against models before and after data removal to show the effectiveness of data removal methods against record-level data leaking. Finally, we evaluate the effect of different levels of amnesiac unlearning on model accuracy, showing the relationship between amount of amnesiac unlearning and model degradation.

\noindent{\bf Experimental Environment:} Our algorithms are implemented in Python 3.7 and use the PyTorch deep learning library~\cite{NEURIPS2019_9015}. All experiments were conducted on the Amazon Sagemaker platform using an ml.g4dn.xlarge instance with 4 vCPUs, 1 GPU, and 16GB of memory. 

\noindent{\bf Datasets:} We conducted experiments on the following two well-known datasets. These datasets were chosen because of the ubiquity of experiments using them as well as to highlight the performance of our algorithm against tasks of varying complexity. 

1. {\bf MNIST handwritten image dataset} \cite{mnist} is a widely used 10-class dataset consisting of 60,000 training images and 10,000 testing images. These images are in grayscale with a resolution of 1x28x28 pixels each.

2. {\bf CIFAR100}~\cite{krizhevsky2009learning} is a 100-class dataset consisting of 600 images from each class. Classes are varied and consist of objects and living things such as dolphins, sunflowers, bottles, and trains. The images have 3 colour channels and have a resolution of 32x32 pixels each.

\noindent{\bf Neural Network:} Our experiments were performed on the Resnet18 convolutional neural network~\cite{he2016deep}, a state-of-the-art residual learning architecture.

\noindent{\bf Attack Algorithms:} Here we describe the two state-of-the-art attack methods considered in our paper.

1. \noindent{\bf Model Inversion Attack:} Our model inversion attack is a slightly modified version of the standard model inversion attack seen in Fredrikson et al.~\cite{Fredrikson2015ModelIA}. The standard attack is deterministic, beginning with a feature vector with all features assigned to 0 (or a suitable starting point for the domain) and labeling this feature vector with the label of the target class $y_t$. The algorithm then performs a forward pass through the model followed by a backward pass, gaining the gradient of loss with regard to the feature vector's classification and $y_t$. Each feature is then shifted in the direction of the gradient, and this process is repeated iteratively, altering the feature vector to be more similar to what the model considers to be an example from that class. 

In the original model inversion attack, a {\it PROCESS} function is performed after each gradient descent step and is intended to help recognition by performing some image processing. We found that this detracted from clarity of generated images, and instead we periodically apply this image processing every $n$ gradient descent steps (where $n \in [500,1000]$). The original attack also begins each inversion from the same state, making each attack deterministic. We found we had better results starting each inversion with a small amount of noise added to each feature, so each inversion is different. Finally, the original attack continues until the change in loss is below some threshold, while we found the attack to be more effective if we continued the attack for some set number of iterations, even while the change in loss was small. These modifications allowed us to generate inversions even on complex convolutional architectures such as Resnet18, a task that was previously judged to be infeasible~\cite{inversiongan}.

2. \noindent{\bf Membership Inference Attack:} We implemented the membership inference attack as described in the following paper~\cite{membership5}. As described, the attack dataset was created by taking the softmax prediction vector $\mathbf{y} = f_{shadow\_model_i}(x)$ from each example $x$ and creating a dataset $\mathit{D}_y$ for each class $y$, where each example is a prediction vector for an example in that class $\mathbf{y}$ and each label is either a 0 or 1, depending on if the example was in the training data for that shadow model. An attack model was then trained on each dataset, and this suite of attack models was used for testing. The attack model is a fully connected network with two hidden layers of width 256 and 128, respectively, with ReLU activation functions and a sigmoid output layer. Increasing the number of shadow models increases the accuracy of the attack but brings additional computational cost.

\subsection{Model Accuracy on Target and Non-target Data}

\begin{figure*}[t!]
\begin{subfigure}{0.33\textwidth}
    \centering
    \includegraphics[width=.95\columnwidth]{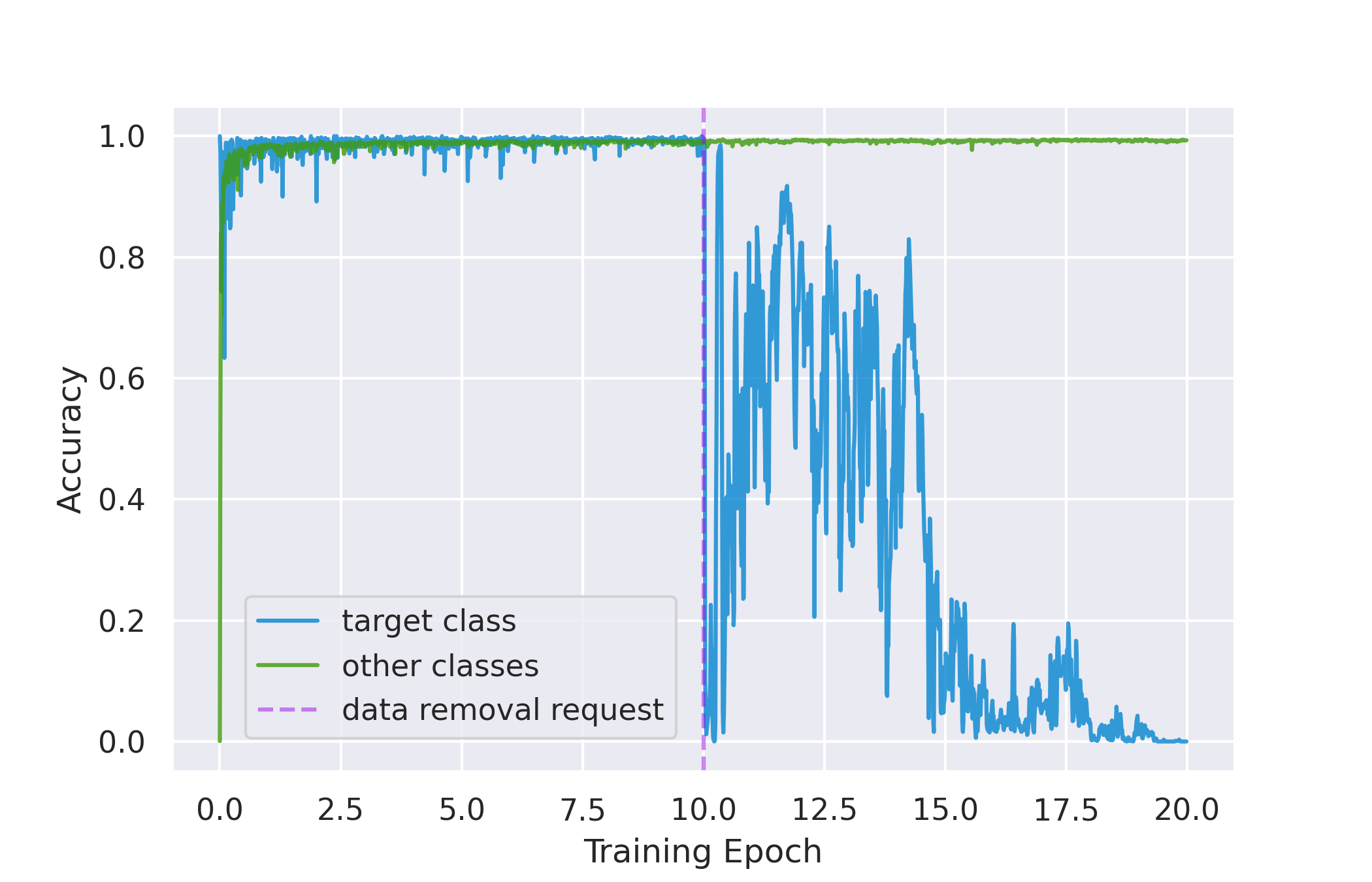}
    \caption{Naive Retraining}
    \label{fig:mnist_naive_resnet}
\end{subfigure}%
\begin{subfigure}{0.33\textwidth}
    \centering
    \includegraphics[width=.95\columnwidth]{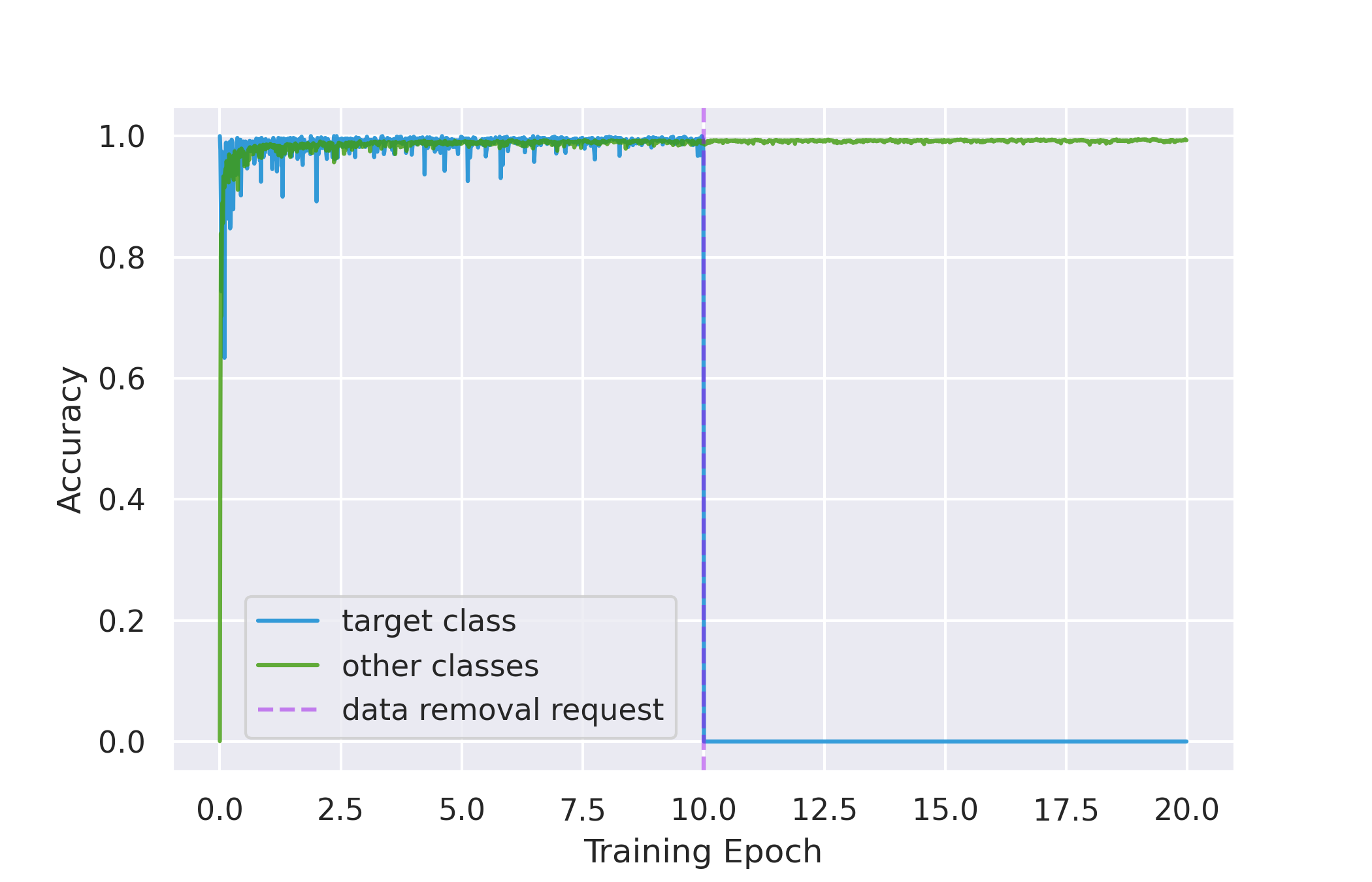}
    \caption{Unlearning}
    \label{fig:mnist_unlearning_resnet}
\end{subfigure}%
\begin{subfigure}{0.33\textwidth}
    \centering
    \includegraphics[width=.95\columnwidth]{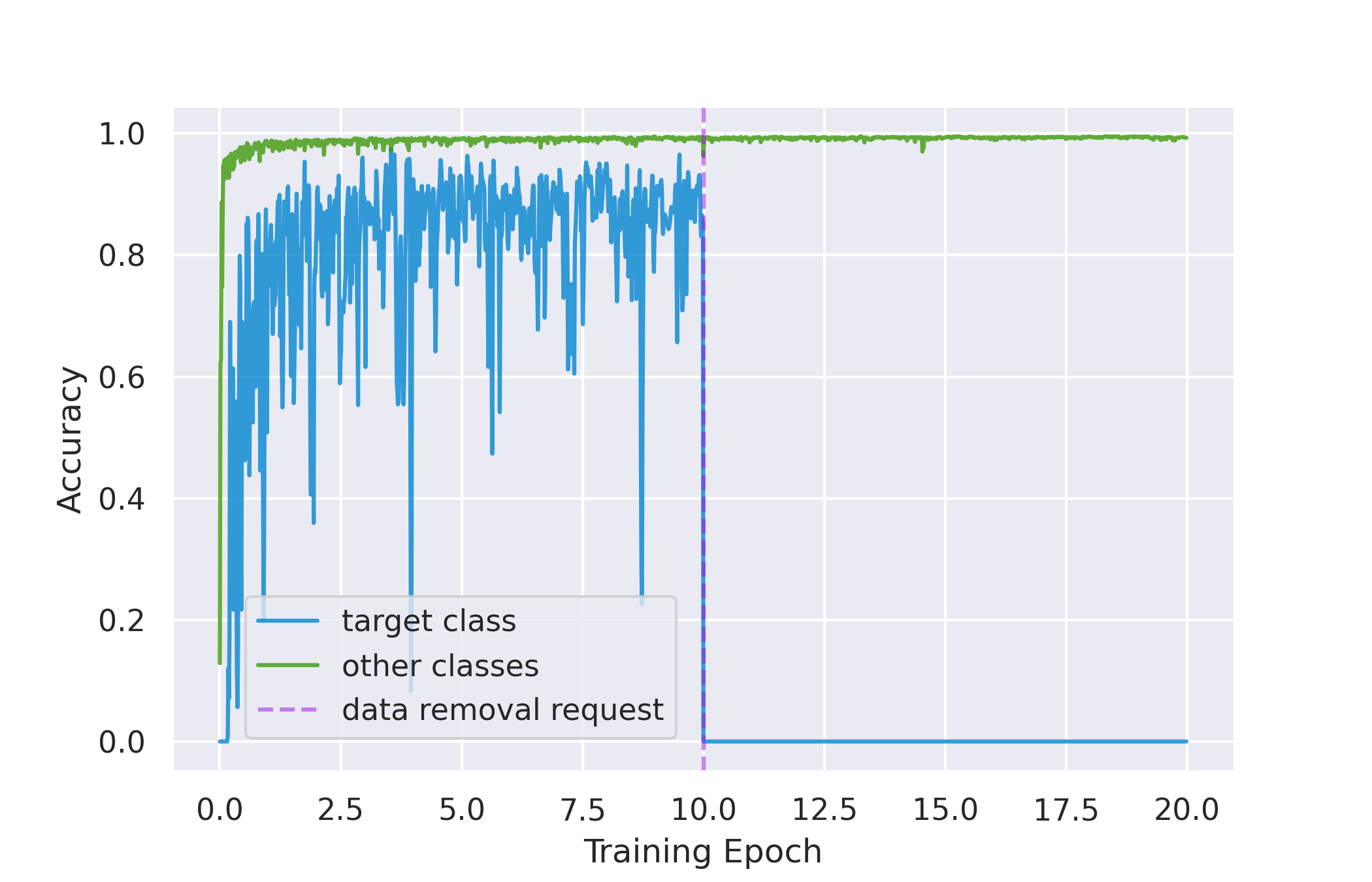}
    \caption{Amnesiac Unlearning}
    \label{fig:mnist_selective_resnet}
\end{subfigure}%
\caption{MNIST model accuracy on target and non-target data}
\end{figure*}

\noindent{\bf MNIST:} Figure~\ref{fig:mnist_naive_resnet} shows the model performance for naive retraining. The model retains a significant amount of knowledge about the sensitive target data for a long period, and the model accuracy on the target data drops slowly over the retraining period. Conversely, figure~\ref{fig:mnist_unlearning_resnet} and figure~\ref{fig:mnist_selective_resnet} show the model accuracy for the sensitive target data drops extremely rapidly for both the unlearning and amnesiac unlearning methods, respectively. In the amnesiac unlearning method, the model accuracy on the non-target data takes a slight dip when the batch learning is initially reversed, but a small amount of training corrects that.

\begin{figure*}[t!]
\begin{subfigure}{0.33\textwidth}
    \centering
    \includegraphics[width=.95\columnwidth]{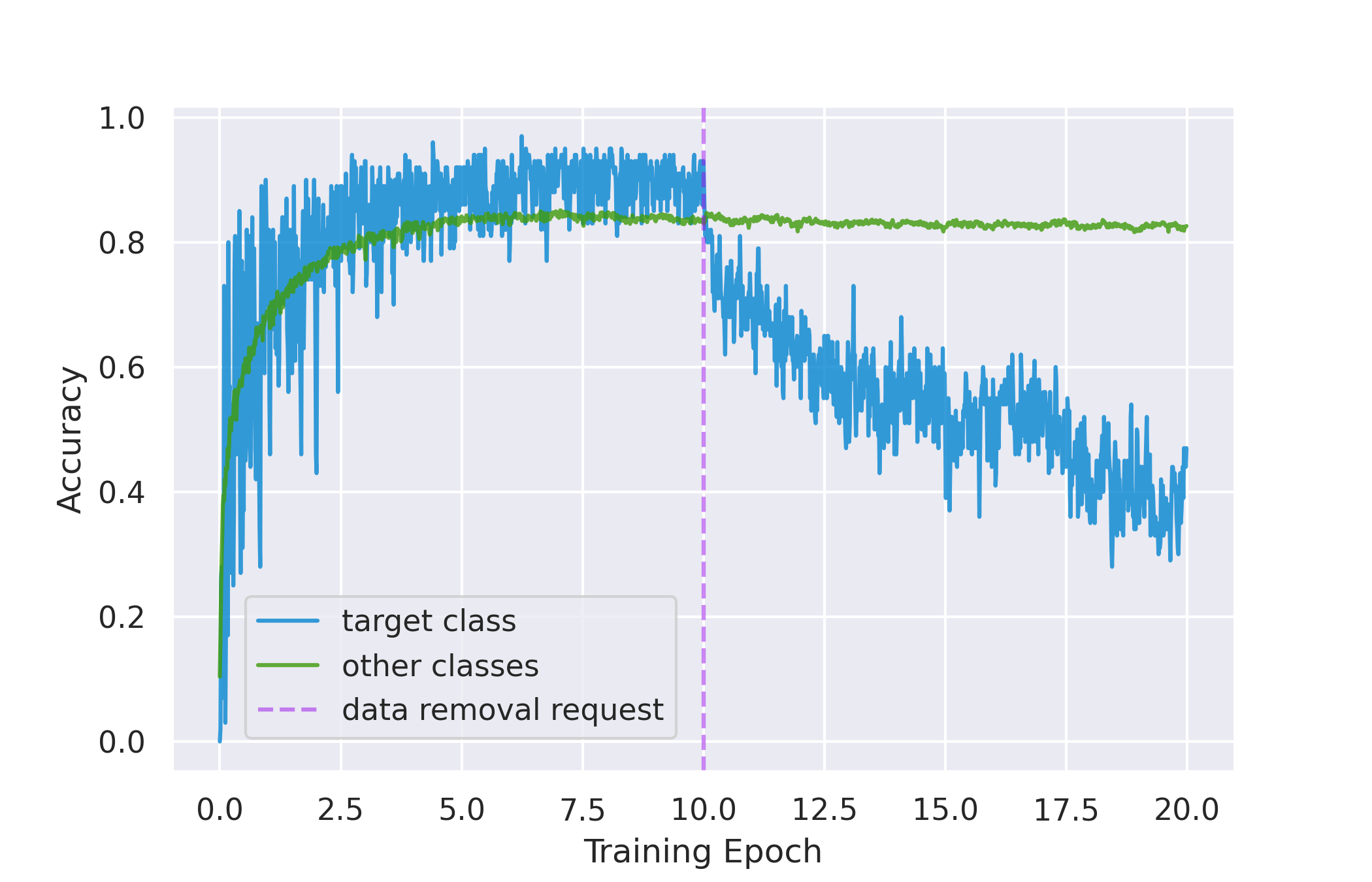}
    \caption{Naive Retraining}
    \label{fig:cifar_naive_resnet}
\end{subfigure}%
\begin{subfigure}{0.33\textwidth}
    \centering
    \includegraphics[width=.95\columnwidth]{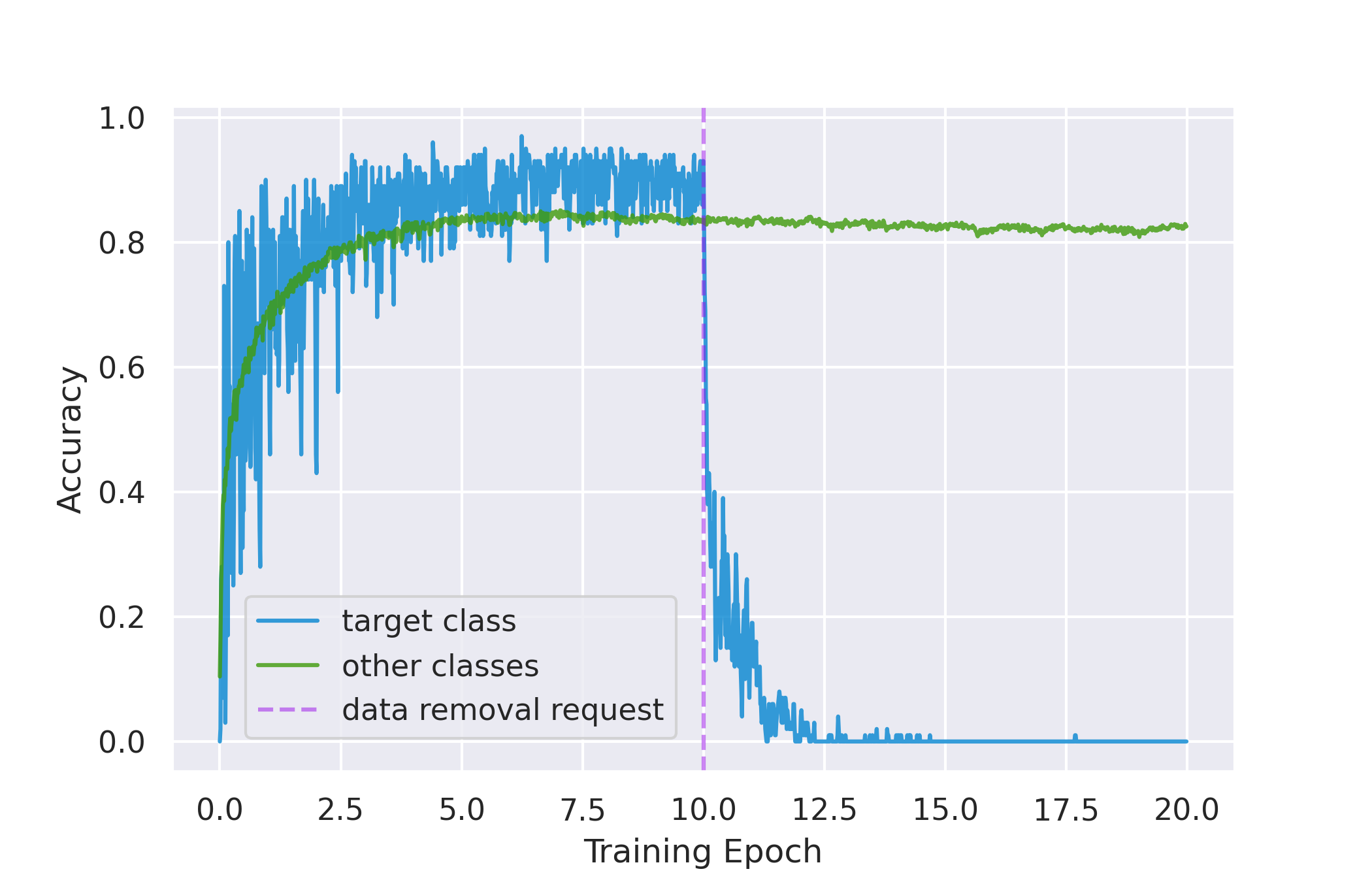}
    \caption{Unlearning}
    \label{fig:cifar_unlearning_resnet}
\end{subfigure}%
\begin{subfigure}{0.33\textwidth}
    \centering
    \includegraphics[width=.95\columnwidth]{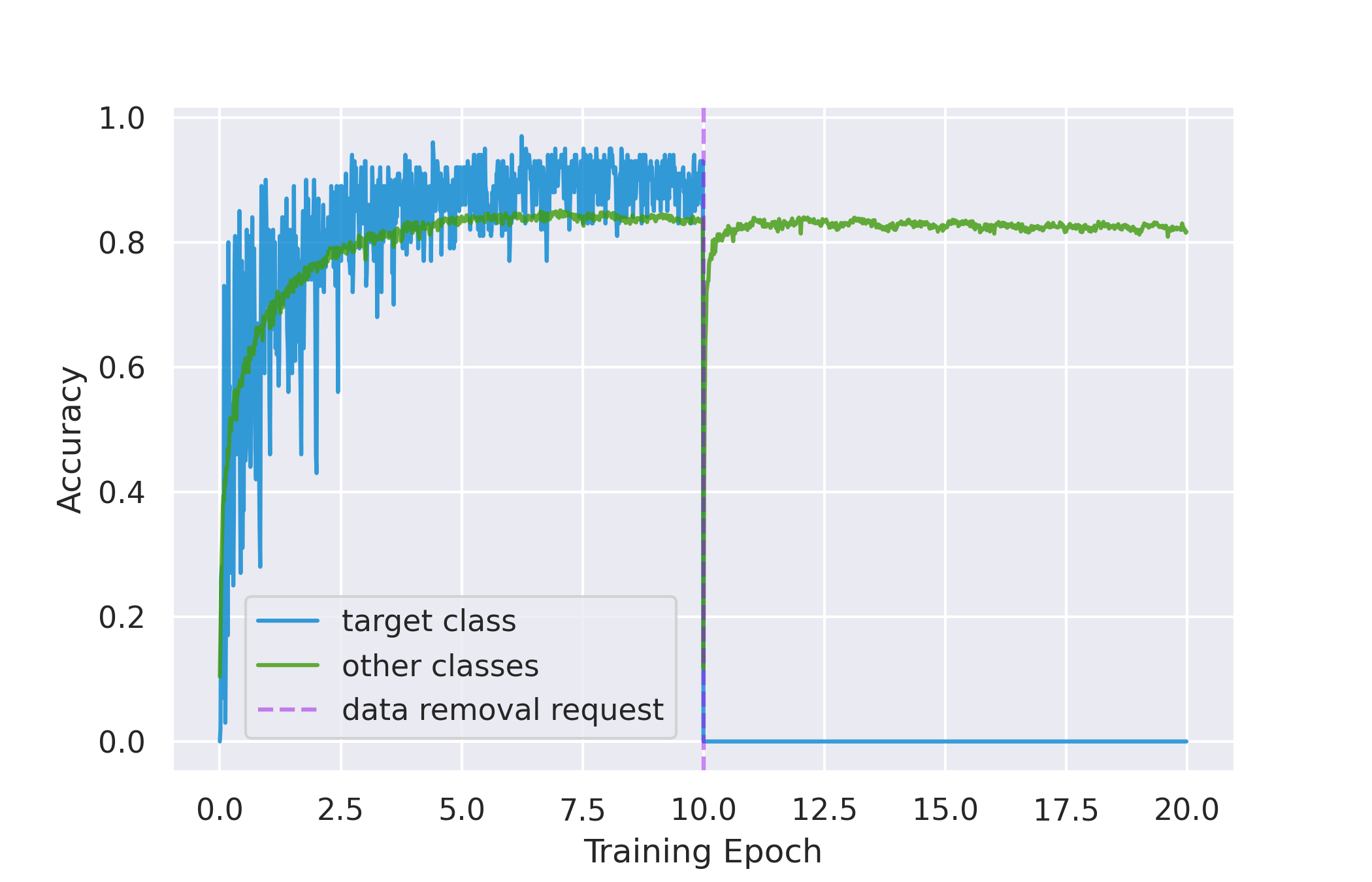}
    \caption{Amnesiac Unlearning}
    \label{fig:cifar_selective_resnet}
\end{subfigure}
\caption{CIFAR100 model accuracy on target and non-target data}
\end{figure*}

\noindent{\bf CIFAR100:} As seen in figure~\ref{fig:cifar_naive_resnet}, the naive retraining method here is even less effective than in the MNIST setting, and after 10 retraining epochs the model still has an approximately 40\% prediction accuracy on the target data. By contrast, the unlearning method (shown in figure~\ref{fig:cifar_unlearning_resnet}) shows a much steeper drop in accuracy on the target data, and within 2 epochs the model has a very small prediction accuracy. After 5 epochs, the unlearning method removes virtually all ability for the model to correctly recognize the target data. The amnesiac unlearning method (shown in figure~\ref{fig:cifar_selective_resnet}) immediately removes the ability for the model to recognize the target data. In both the naive and unlearning methods the prediction accuracy of the non-target data stays high, showing that these methods do not degrade the learned information for non-target data. The amnesiac unlearning method shows an initial accuracy loss when the batches are reversed, but this quickly rebounds after a small amount of training. In this instance, approximately 6\% of the batches were removed, which makes for a significant effect on the rest of the model. This effect is examined in greater detail in the {\bf Effect of Amnesiac Unlearning on Model Efficacy} section.

\subsection{Model Inversion Attacks}

\begin{figure}[t]
    \centering
    \includegraphics[width=.5\columnwidth]{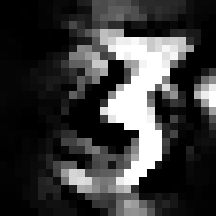}
    \caption{Model Inversion Attack on trained model}
    \label{fig:inversion}
\end{figure}

Model inversion attacks were performed on trained MNIST models both before and after the data removal methods were applied. Each model started from the same before state, and an inversion attack against the trained model can be seen in figure~\ref{fig:inversion}. Further inversion attacks were performed during and after the data removal process. In the case of the naive retraining and unlearning methods, inversion attacks were performed after each epoch of training on the dataset modified to cause data removal. In the case of amnesiac unlearning, the amnesiac unlearning was performed for all batches containing the target class and then inversion attacks were performed after each epoch of training on a modified dataset without the target class. All inversion were performed targeting the class representing the digit {\bf 3}, and all data removal methods likewise attempted to remove knowledge of that class. Due to the stochastic nature of our modified inversion attack, we performed multiple attacks against each model and selected the ones that most resemble the target class. Unfortunately, it is very difficult to come up with an acceptable metric to quantify the "recognizability" of inverted images, and visual recognition was our best way of judging the success of the attacks.

\begin{figure}[t]
\centering

\begin{tabular}{ccc}
    \includegraphics[width=0.3\columnwidth]{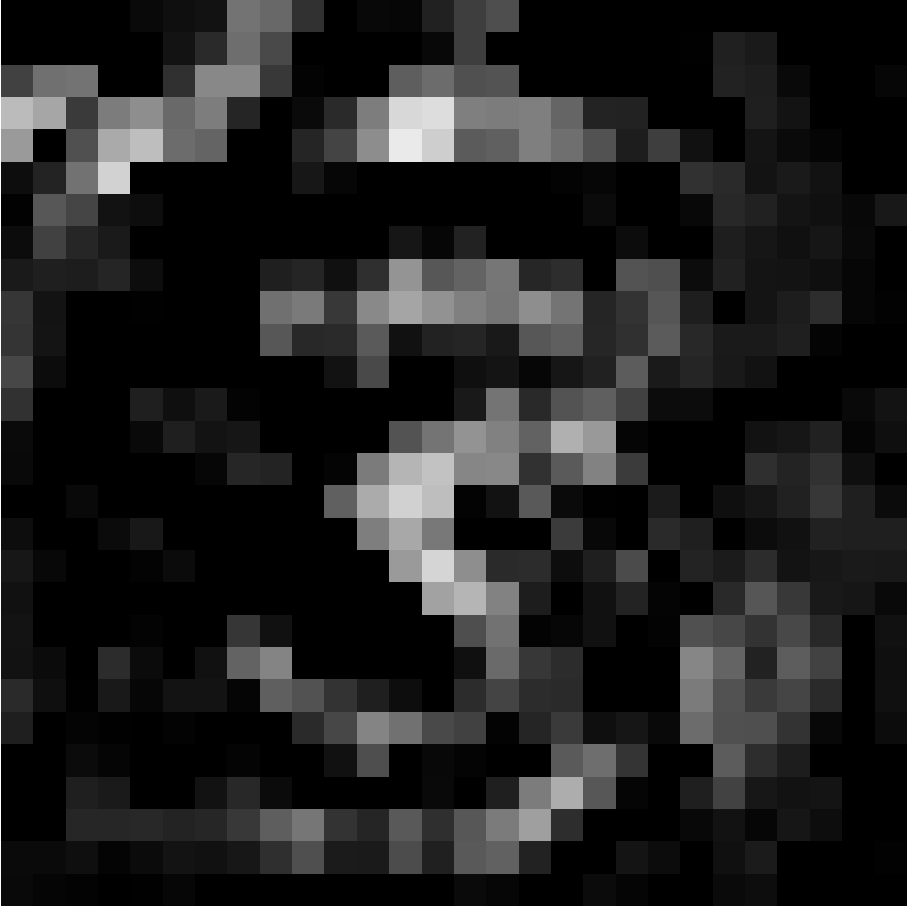} & 
    \includegraphics[width=0.3\columnwidth]{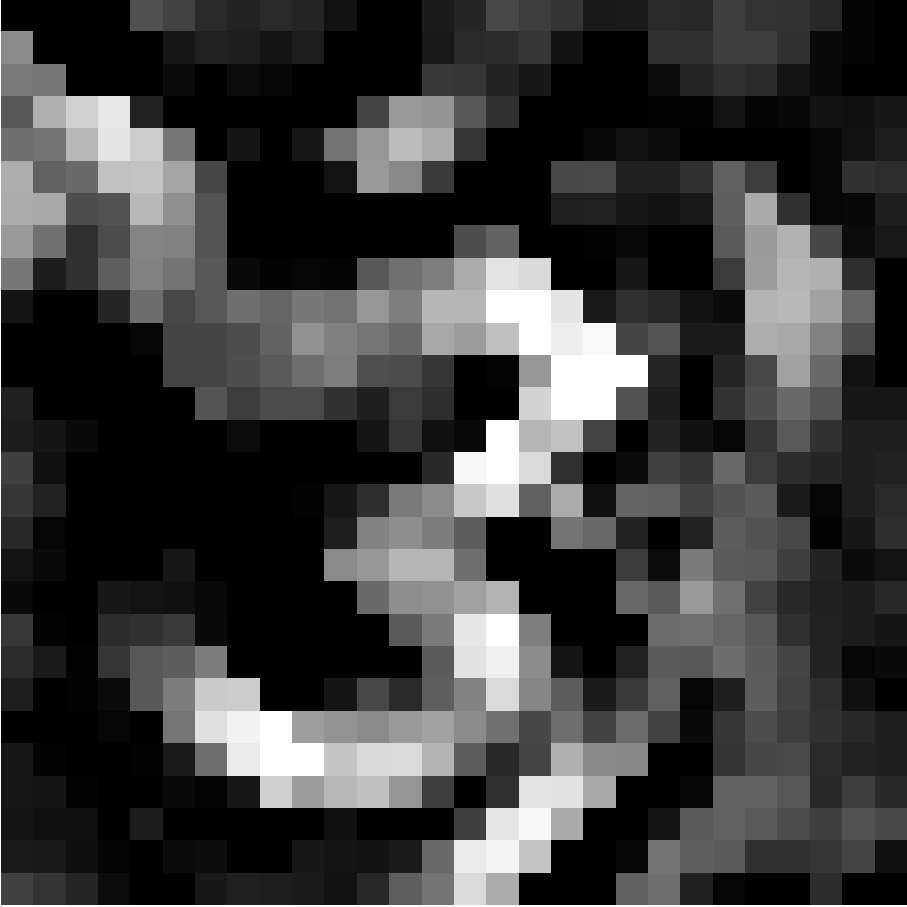} &
    \includegraphics[width=0.3\columnwidth]{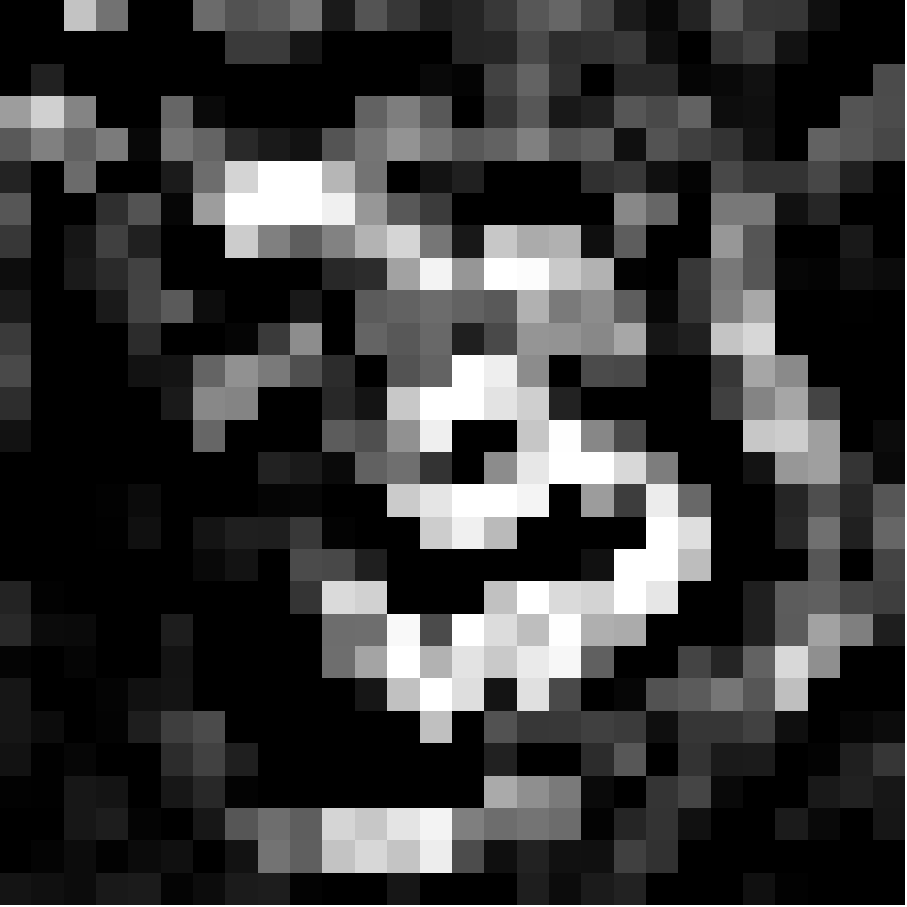}\\
     & 
\end{tabular}
\caption{Model Inversion Attack results after 1, 5, and 10 epochs of naive retraining}
\label{fig:inv_naive}

\end{figure}

\begin{figure}[t]
\centering

\begin{tabular}{ccc}
    \includegraphics[width=0.3\columnwidth]{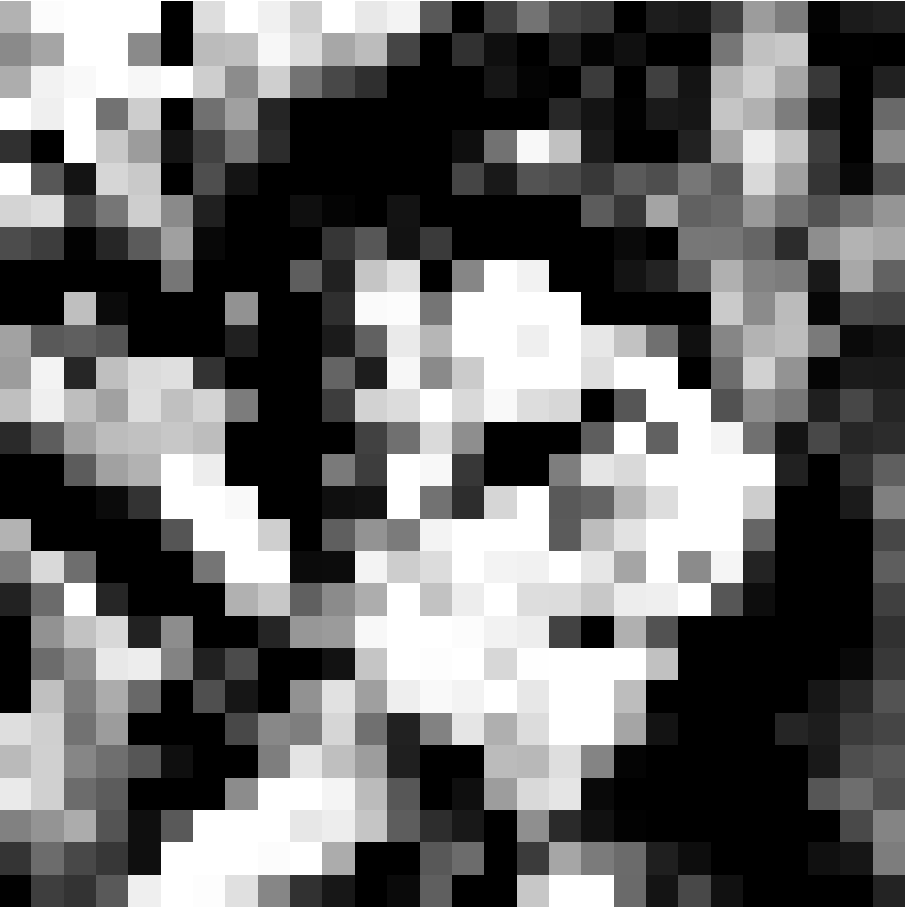} & 
    \includegraphics[width=0.3\columnwidth]{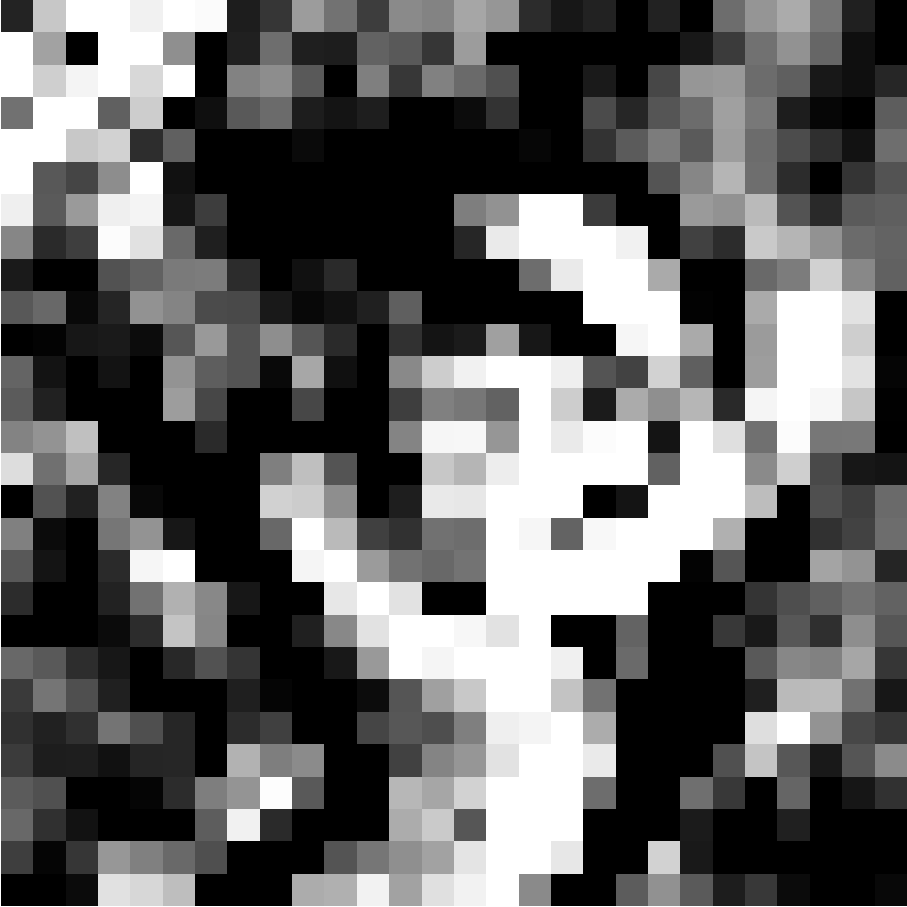} &
    \includegraphics[width=0.3\columnwidth]{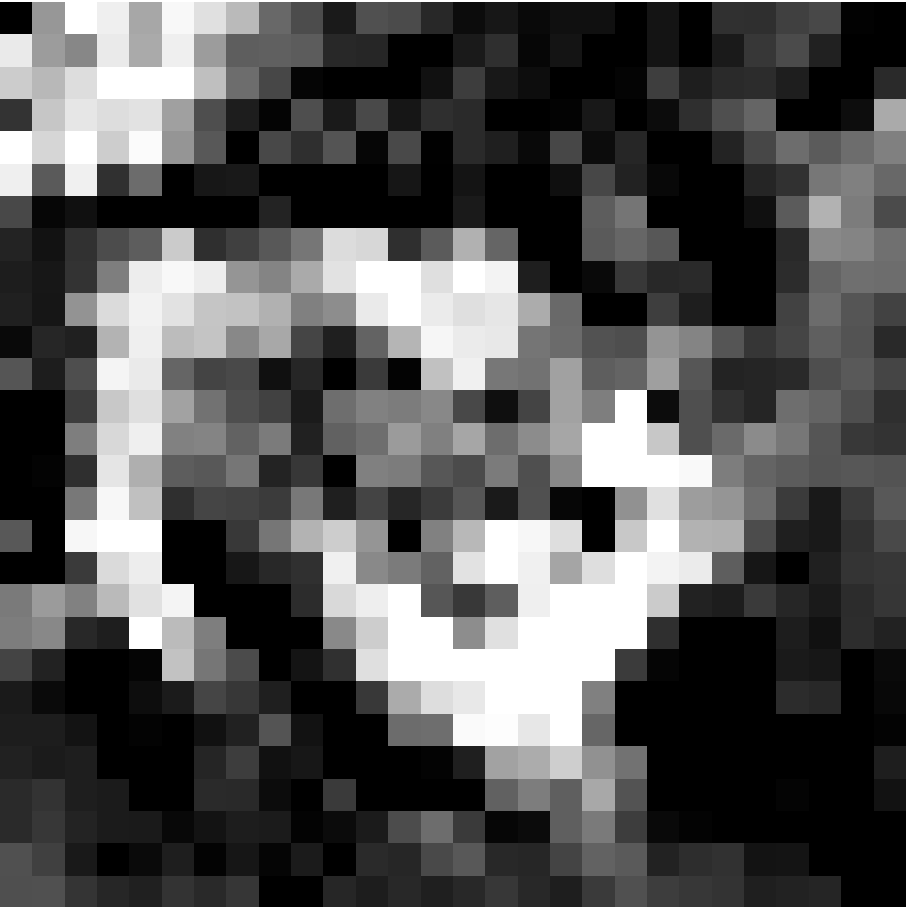}\\
     & 
\end{tabular}
\caption{Model Inversion Attack results after 1, 5, and 10 epochs of unlearning}
\label{fig:inv_unlearning}

\end{figure}

\begin{figure}[t]
\centering

\begin{tabular}{ccc}
    \includegraphics[width=0.3\columnwidth]{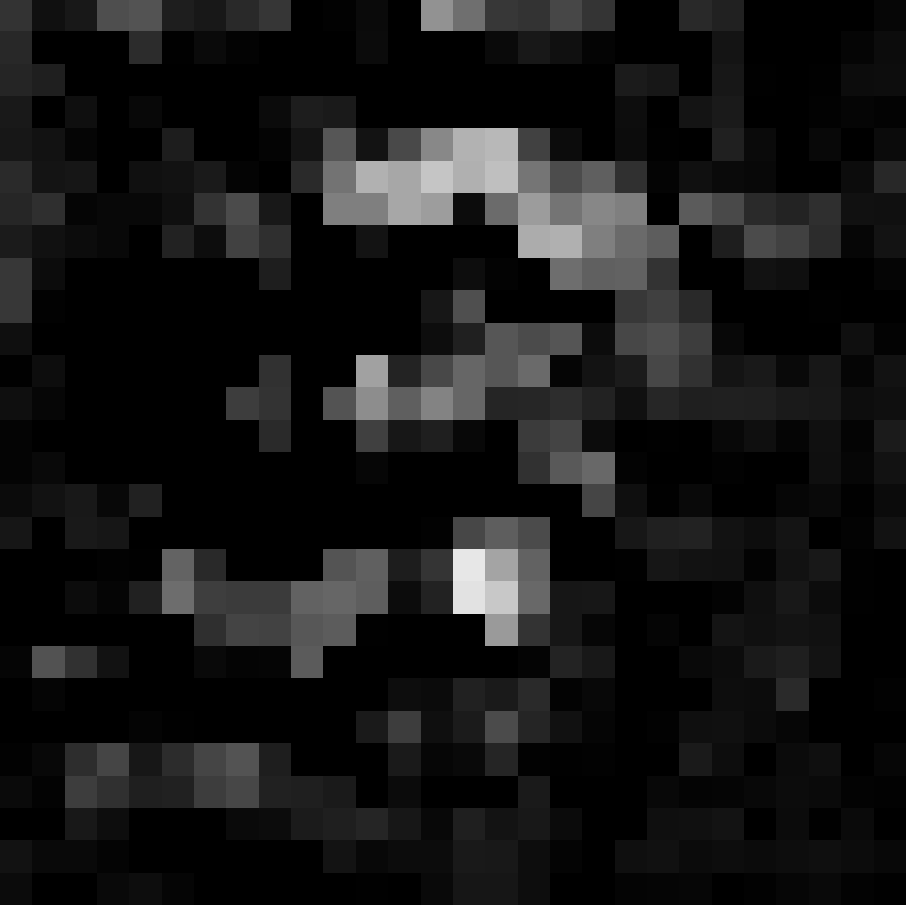} & 
    \includegraphics[width=0.3\columnwidth]{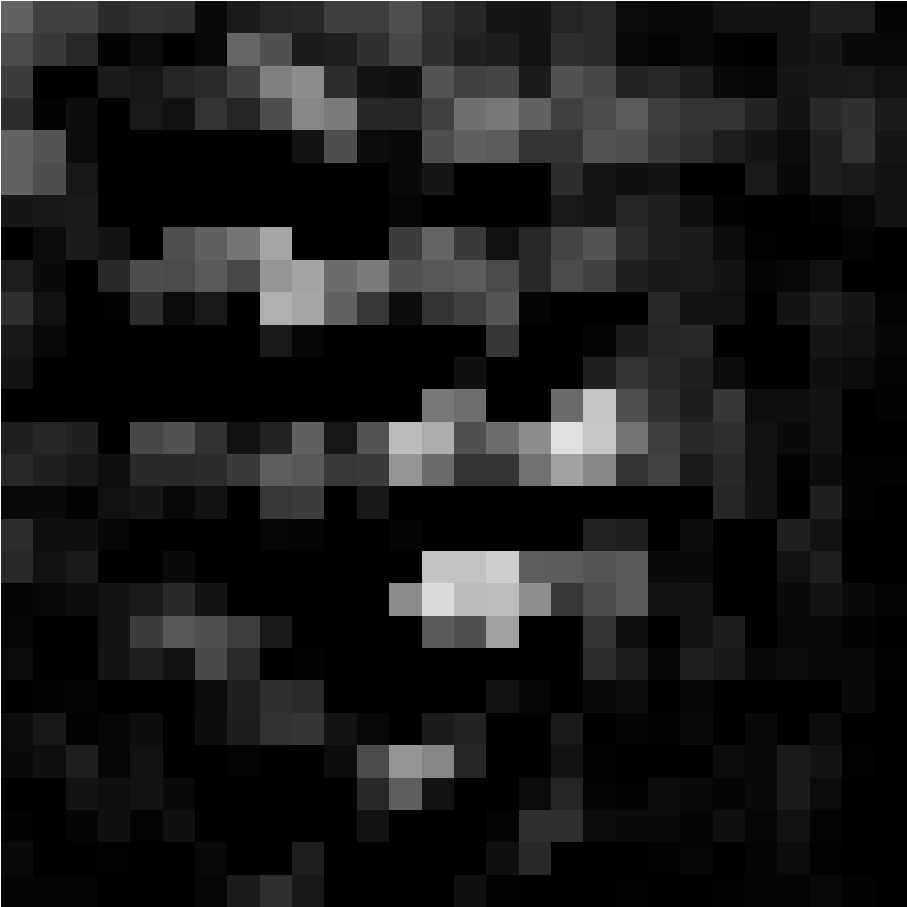} &
    \includegraphics[width=0.3\columnwidth]{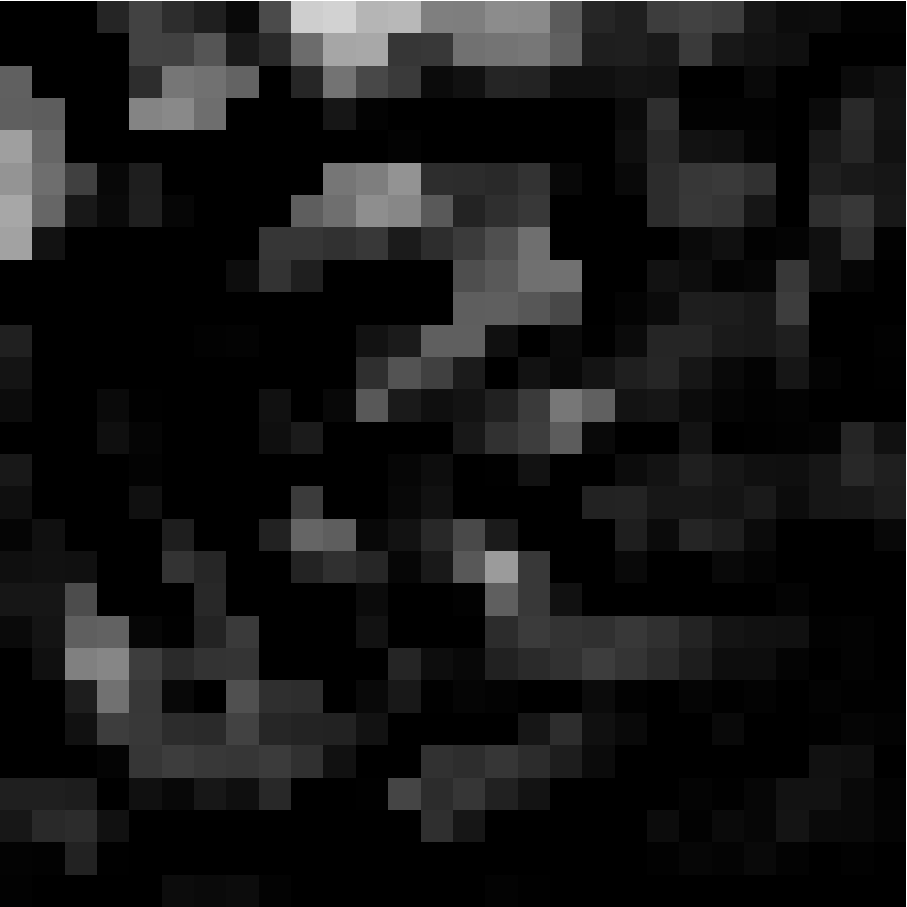}\\
     & 
\end{tabular}
\caption{Model Inversion Attack results after 1, 5, and 10 epochs after amnesiac unlearning}
\label{fig:inv_selective}

\end{figure}

As seen in figure~\ref{fig:inv_naive}, the naive retraining method does very little to protect against leaking private class information. Even after 10 retraining epochs, inversions targeting the class are still somewhat recognizable. In conjunction with the test accuracy results, this underlines the unsuitability of naive retraining to protect private information.

The model inversion attacks results against the models effected by unlearning can be seen in figure~\ref{fig:inv_unlearning}. This method almost immediately completely removes any ability to gain useful class information with model inversion attacks, showing how effective this method is at protecting against this sort of data leaking attack.

For the amnesiac unlearning method, the model inversion attacks have remarkably little gradient information to go off of, and as a result the images (seen in figure~\ref{fig:inv_selective}) are dark and jumbled, although they do seem to hint toward the general shape of the target data. As expected, more training on the modified dataset removes the ability to gain any meaningful information about the target data, and the attacks after 5 and 10 epochs of training are almost unrecognizable. This suggests that this method is best used in situations where a small number of data points needs to be removed, as opposed to an entire class of data (in this case, a tenth of learned data).

\subsection{Membership Inference Attacks}

Membership inference attacks were performed with 16 shadow models, each trained on CIFAR100 for 10 epochs. The target model was likewise trained for 10 epochs. We evaluate the effectiveness of membership inference attacks using the recall metric, a metric that give us great insight into how effective these attacks are at leaking data. This metric is more helpful than accuracy in an environment where we care a lot about correctly recognizing positive instances. This gives us information about how effective our method is at preventing membership inference data leaks. All membership inference attacks were performed targeting a set of individual examples, and then the data removal techniques were performed to attempt to remove learned data from this set of individual examples.

Membership inference attacks were performed on a trained model, and the recall value of this attack can be seen in table~\ref{tab:inference} at epoch 0. Data removal techniques were then performed, and the result of membership inference attacks on the model after the application of the removal technique but before any retraining can be seen at epoch 0'. In the case of naive retraining and unlearning, this is simply changing the datasets and has not had an effect, while in the case of amnesiac unlearning it represents the reversal of the batches containing the sensitive data. Subsequent membership inference attacks were performed after each epoch of training on the modified datasets.

\begin{table}[t]
    \centering
    \begin{tabular}{|p{.2\columnwidth}|p{.2\columnwidth}|p{.2\columnwidth}|p{.2\columnwidth}|}
    \hline
        Epoch & Naive Retraining & Unlearning & Amnesiac Unlearning\\
        \hline
        0 & 0.97478991 & 0.97478991 & 0.97478991 \\
        0' & 0.97478991 & 0.97478991 & 0.0 \\
        1 & 0.09243697 & 0.0 & 0.0 \\
        2 & 0.05882352 & 0.0 & 0.0 \\
        3 & 0.0 & 0.0 & 0.0 \\
        4 & 0.0 & 0.0 & 0.0 \\
        5 & 0.0 & 0.0 & 0.0\\ \hline
    \end{tabular}
    \caption{Recall of membership inference attacks}
    \label{tab:inference}
\end{table}

The results shown in table~\ref{tab:inference} show that the naive retraining method does not prevent the membership inference attacks for more than 2 full epochs of retraining. This highlights the insufficiency of this method to protect against data leaking attacks, and emphasizes the need for other methods. By contrast, the unlearning and amnesiac unlearning methods both protect against membership inference attacks with less than an epoch of retraining (and in the case of amnesiac unlearning, with no retraining whatsoever). 
\subsection{Effect of Amnesiac Unlearning on Model Efficacy}

\begin{figure}[t]
    \centering
    \includegraphics[width=.95\columnwidth]{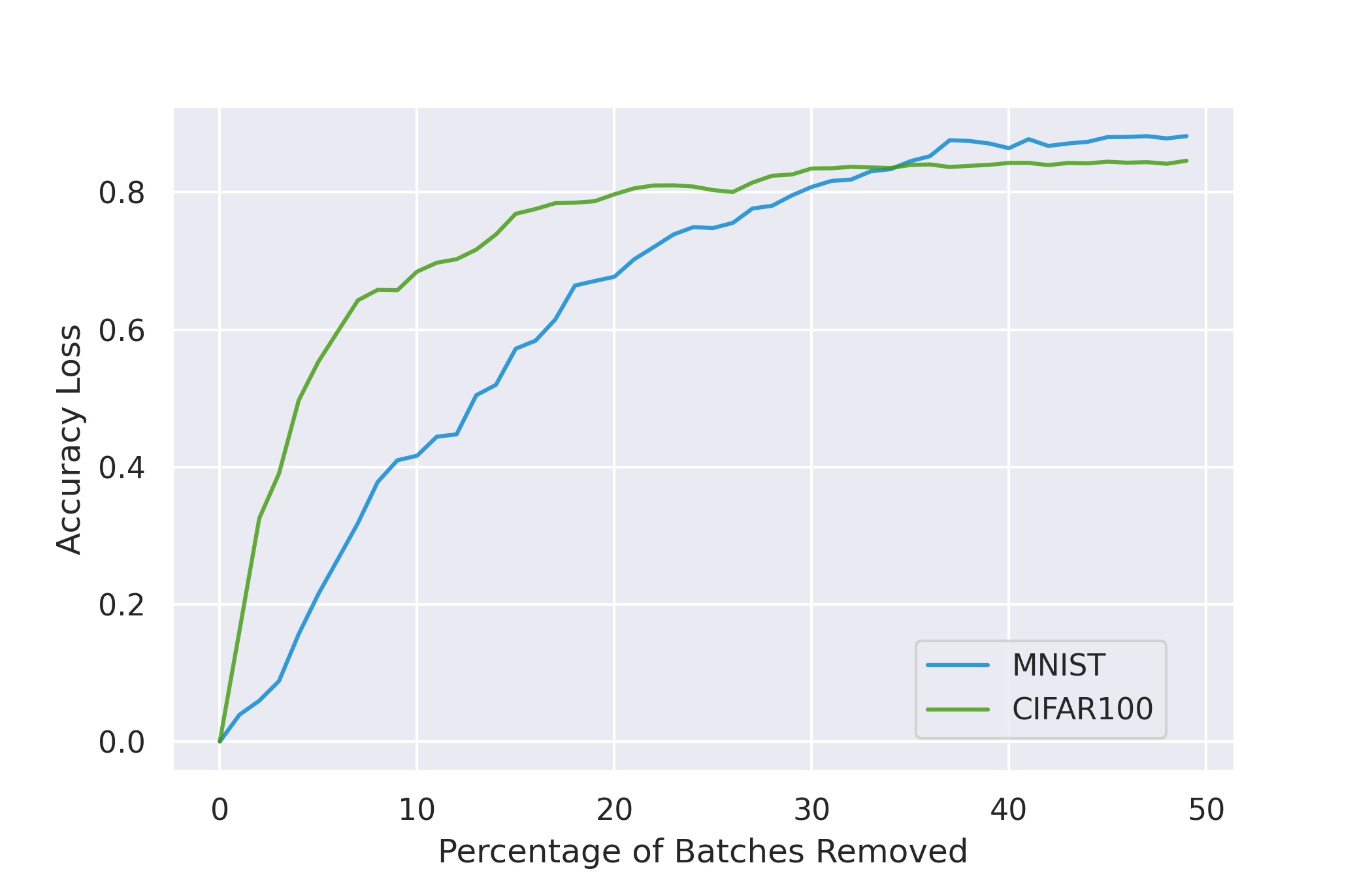}
    \caption{Test accuracy on all classes with increased amnesiac unlearning (mean of 10 runs)}
    \label{fig:amnesiac_mnist}
\end{figure}

Amnesiac unlearning, while extremely effective at removing the learning from specific data examples, has a significant effect on the efficacy of the model if it is used too often. With 1\% or less of the batches removed the model accuracy remains close to what it was before any amnesiac unlearning was performed, but this quickly drops as the learning from more batches is removed, as seen in figure~\ref{fig:amnesiac_mnist}. Drops in accuracy can be easily remedied with a small amount of fine-tuning, but this underscores the need for the user to be wise in selecting their data removal method - if they have a small amount of data that needs to be removed, amnesiac unlearning is efficient and effective, without significantly effecting the rest of the model. However, if a larger amount of data needs to be removed, consideration should be given to unlearning, which is able to remove larger amounts of learned data without needing the same fine-tuning after the removal step.

\section{Discussion}

\noindent{\bf Metrics:} We emphasize that while test accuracy, model inversion attacks, and membership inference attacks can give useful information about a model's propensity to leak private information about sensitive data, they are not methods of measuring how much information has been retained about that sensitive data. Some models and settings aren't vulnerable to model inversion attacks and membership inference attacks can be protected against with methods such as differential privacy, but neither of these things provide a guarantee that models cannot leak other private information. However, finding a comprehensive method of evaluating how much private data can be leaked from a model through any means is a difficult task, and to-date we are not aware of any method that claims to do so.

\noindent{\bf Black-box Attacks:} One may argue that the impact of model inversion and membership inference attacks can be mitigated by allowing only black-box access to the models themselves. Unfortunately, model extraction attacks~\cite{extraction1},~\cite{extraction2},~\cite{extraction3} have the ability to steal functionality of models even with only black-box access, and in some cases even with only access to truncated prediction vectors~\cite{extraction4}. The existence and success of these attacks show that limiting access to black-box access alone is not sufficient to protect against motivated attackers who can steal a model and then attack it under a white-box setting. 

\section{Conclusion}

In this paper, we examined and evaluated methods aimed at removing learned data from trained neural network models. This is especially relevant because of right to be forgotten regulations, such as in the European Union's GDPR law, that require data holders to delete data on individuals when requested. Training new models from scratch can be prohibitive, and there is an acute need for methods that can effectively and efficiently remove learned data from the models to protect user privacy, while at the same time preserving model performance on non-target data. We introduce two novel methods of data removal, namely {\it unlearning} and {\it amnesiac unlearning}, that can be used to protect privacy of target (sensitive) data without incurring significant cost or degrading model performance on non-target data. We evaluated these methods against two state-of-the-art data leaking attacks, namely model inversion and membership inference, and showed that both data removal methods are effective at protecting data from leaking. An additional interesting finding was that the unlearning method is better for removing large amounts of learned data, while amnesiac unlearning is better for laser-focused removal of specific sections of data, such as a single example or set of examples. Given that this line of research at the intersection of data privacy and machine learning is a relatively new and very rich field, there remains a number of important problems to be solved, including a method of measuring data retention after data removal techniques are applied.

\bibliography{amnesiacml}

\end{document}